# A Framework for Evaluating Agricultural Ontologies

A. Goldstein [1*], L. Fink[2] and G. Ravid[2]

[1] Tel-Aviv University, Department of Technology and Information Management, Israel

[2] Ben Gurion University of the Negev, Department of Industrial Engineering and Management, Israel

## Abstract

*An ontology is a formal representation of domain knowledge, which can be interpreted by machines. In recent years, ontologies have become a major tool for domain knowledge representation and a core component of many knowledge management systems, decision support systems and other intelligent systems, inter alia, in the context of agriculture.*

*A review of the existing literature on agricultural ontologies, however, reveals that most of the studies, which propose agricultural ontologies, are lacking an explicit evaluation procedure. This is undesired because without well-structured evaluation processes, it is difficult to consider the value of ontologies to research and practice. Moreover, it is difficult to rely on such ontologies and share them on the Semantic Web or between semantic aware applications. With the growing number of ontology-based agricultural systems and the increasing popularity of the Semantic Web, it becomes essential that such development and evaluation methods are put forward to guide future efforts of ontology development.*

*Our work contributes to the literature on agricultural ontologies, by presenting a method for evaluating agricultural ontologies, which seems to be missing from most existing studies on agricultural ontologies. The framework supports the matching of appropriate evaluation methods for a given ontology based on the ontology's purpose.*

**Introduction**

An ontology is a formal representation of domain knowledge, which can be interpreted by machines (Chandrasekaran *et al.* 1999). In other words, ontologies formally define the entities (concepts) of a domain, their attributes and the relationships among them, in a machine interpretable way. Thus, ontologies have become a major tool for domain knowledge representation and a core component of many knowledge management systems, decision support systems and other intelligent systems (Bose and Sugumarat 2007; Bose and Sugumarat 2007; Chandrasekaran *et al.* 1999; Delir Haghighi *et al.* 2013; Yu *et al.* 2005; Noy and McGuinness 2001).

Ontologies can be used for several purposes: First, an ontology, being a formal explicit description of domain knowledge, can be used by machines for knowledge deduction (W3C 2006). For example, if a data set specifies that a certain chemical "x" is effective against a certain pest "y" and that a certain pesticide "z" contains that chemical "x", then a machine can infer that the pesticide "z" is effective against this pest "y". Such inference can be made if an ontology declares that a pesticide is effective against a pest if a chemical it contains is effective against the pest. This makes ontologies suitable for serving as the underlying knowledge base of decision support systems (DSS) and expert systems.

Second, ontologies enable sharing conceptual schemata of data, allowing application programs and databases to interoperate without having to share data structures (Gruber 1993; W3C 2006). As a result, if two application programs or webpages share an ontology then it is possible to automatically extract and aggregate information from these applications and webpages. In addition, it is possible to translate concepts of one application to concepts representing the same thing in the other application.

Third, ontologies enable reuse of domain knowledge (Noy and McGuinness 2001; Roussey *et al.* 2010). Once an ontology is published it can be used by various applications in various domains. Furthermore, several ontologies describing different parts of a domain can be integrated to create one large ontology of the domain (Noy and McGuinness 2001).

A pivotal use of ontologies is for the creation of the so called *Semantic Web*. The Semantic Web (also known as *Web of Data* or *Web* of *Linked Data*) is an extension of the Web through standards by the World Wide Web Consortium (W3C) with the aim of providing a formal representation of the information on the World Wide Web, to facilitate sharing and reusing of data on the web across applications (Berners-Lee *et al.* 2001; W3C 2006).

Currently, most of the data on the web reside in HTML documents that can be read by humans and by machines. While humans are capable of understanding the meaning of these documents, machines cannot extract meaning from these documents other than searching them for particular keywords. The Semantic Web initiative is aimed at changing this situation and making data in Web documents understandable for machines – by using ontologies. Ontologies provide a formal language that specifies how data on the web is related to objects (i.e. instances of classes) in the real world. In addition, since these ontologies are described in standard formats it is possible to integrate and combine data from diverse sources on the Web, thereby making the Web one huge database.

To support the Semantic Web initiative, the W3C has developed a set of standards and tools. For example:

- The *Resource Description Framework (RDF)* – a formal language for specifying relations between web resources that represent objects in the real world using triples in the form of subject-predicate-object.
- The RDF Schema (RDFS) – an extension of RDF with additional classes that allow the specification of the ontology's schema or data-model, thereby adding semantics to ontologies.
- The Web Ontology language (OWL) – an extension of RDF, which allows the

specification of ontologies with higher level of semantics that include operations on classes (e.g. union and intersection) and is able to express description logic predicates, which can be used for inference and assertions. OWL also allows binding a concept of one ontology with a similar concept of another ontology (using "owl:sameAs" property), thereby making additional concepts of the other ontology available for use, even though defined in a different ontology.

- The Simple Knowledge Organization System (SKOS) – a standard which is built upon RDF and RDFS and is used for the specification and publication of vocabularies on the Semantic Web.

- SPARQL – a standardized query language, which enables querying decentralized collections of RDF data that are stored in one or more *triplestores*. A triplestore is a special database for the storage and retrieval of RDF triples.

The increasing use of ontologies is also seen in agriculture (e.g.Beck *et al.* 2010; Chang *et al.* 2008; Kragt *et al.* 2016; Li *et al.* 2013; Liao *et al.* 2015; Roussey *et al.* 2010; Song *et al.* 2012; Tomic *et al.* 2015; Zhang *et al.* 2002), where they are used for various purposes, such as agriculture knowledge sharing across farmers around the world (and in different languages) (AGROVOC Thesaurus; Chang *et al.* 2008; Maliappis 2009), creating semantic interoperability of agricultural systems (Aqeel-ur and Zubair 2011; Goumopoulos *et al.* 2009; Tomic *et al.* 2015), and supporting farmer decisions (Gaire *et al.* 2013). This is not surprising, given that agriculture is a knowledge-centric field that covers many areas of expertise, many world-wide used practices and technologies, and includes numerous concepts that are often designated by different names with similar meaning (Liao *et al.* 2015; Palavitsinis and Manouselis 2014) and fragmented across different systems (Janssen *et al.* 2017).

A review of the existing literature on agricultural ontologies, however, reveals that most of the studies, which propose agricultural ontologies, are lacking a clear ontology construction method and, more importantly, explicit evaluation procedures. This is undesired because without well-structured development and evaluation processes, it is difficult to consider the value of ontologies to research and practice. Moreover, it is difficult to rely on such ontologies and share them on the Semantic Web or between semantic aware applications. **Error! Reference source not found.** summarizes for each of the surveyed agricultural ontologies, their goal and whether their construction and evaluation processes are available.

With the growing number of ontology-based agricultural systems and the increasing popularity of the Semantic Web, it becomes essential that such development and evaluation methods are put forward to guide future efforts of ontology development.

**Table 1.** A summary of agricultural ontologies, their goals, and whether their construction and evaluation processes are described

| Goal \ Study | Share vocabularies, integrate data | Knowledge search and exploration | System interoperability | Decision support | Construction method described | Evaluation method described |
|---|---|---|---|---|---|---|
| (Roussey et al. 2010) | ✓ | ✓ | ✓ | ✓ | - | - |
| (Chang et al. 2008) | ✓ | ✓ | | | - | - |
| (Goumopoulos et al. 2009) | | | ✓ | | + | + |
| (Aqeel-ur and Zubair 2011) | | | ✓ | | - | - |
| (Gaire et al. 2013) | | | ✓ | ✓ | - | - |
| (Tomic et al. 2015) | ✓ | | | | + | - |
| (Li et al. 2013) | | ✓ | | | + | - |
| (Song et al. 2012). | | ✓ | | | - | - |

| (Liao et al. 2015) | ✓ |   | ✓ |   | Partial | + |
| (Beck et al. 2005) |   | ✓ |   |   | + | - |
| (Maliappis 2009) | ✓ |   | ✓ |   | Partial | - |

Our work extends the existing literature on agricultural ontologies, by presenting a comprehensive method for evaluating agricultural ontologies. The method is demonstrated on the case of the pest-control ontology.

The rest of the paper is organized as follows: next, in the Materials and methods section, we present a review of ontology evaluation methods, as well as survey the purposes for which ontologies are used in the context of agriculture. In the Results section, we present the framework for matching evaluation methods based on the ontology purpose. We demonstrate how the framework should be applied in a case study of pest-control ontology. Finally, we discuss the resulting framework and provide conclusion.

**Materials and methods**

The development of the evaluation framework for agricultural ontologies included three steps: First, literature on existing ontology evaluation approaches (not necessarily in the context of agriculture) are surveyed.

Second, given the goals / uses of ontologies in the context of agriculture, which were identified in the Introduction section, we define a framework that matches evaluation approaches to different ontology purposes.

Third, we demonstrate the application of the framework, using a case-study of a pest-control ontology, which has been developed for integrating knowledge from different websites as well as concepts from other ontologies in order to provide a pest-control knowledge base, and for supporting pest-control decisions of farmers, as it serves as the knowledge base of a pest-control decision support application.

*A Review of Ontology Evaluation Methods*

Various ontology evaluation methods have been proposed in the literature. These methods are commonly classified into the following types (Brank *et al.* 2005; Yu *et al.* 2007):

- *Evaluation against a gold standard* - This method compares an ontology with common standards or with another ontology that is considered as a benchmark. Such a method is typically used in cases where the ontology was automatically or semi-automatically generated. In many cases the application of this method is impossible since such a gold standard does not exists.

- *Application-based evaluation* - The application based evaluation (or task-based evaluation Yu *et al.* 2007) uses the ontology for completing tasks within an application and measure its effectiveness. Since comparing several optional ontologies in the context of a given application environment is usually not feasible, often the proposed ontology is evaluated in a quantitative or qualitative manner by measuring its suitability for performing tasks within the application. The advantage of this method is that it allows assessing how well the ontology fulfills its objectives. Nevertheless, the evaluation is only relevant for that particular application. If the ontology is to be used in another application the evaluation is irrelevant.

- *Criteria-based evaluation* - This method evaluates the ontology against a set of predefined criteria. Depending on the criteria, the evaluation is conducted either automatically (Yu *et al.* 2007) or manually, usually by experts (Delir Haghighi *et al.* 2013). Various criteria have been used in the literature. Gruber (1995) defines five criteria: *Clarity* – the ontology should effectively and objectively communicate the definitions of terms; *Coherence* – the ontology should support inferences that are consistent with the definitions and have no contradictions; *Extendibility* – it should be possible to extend the ontology to support possible uses of the shared vocabulary, without altering the ontology; *Minimal encoding bias* – the conceptualization should be independent of the particular encoding that is used as possible; and *Minimal ontological commitment* – the ontology should define as few restrictions on the domain of discourse as possible.

Gómez-Pérez (1996) also defines five criteria that are partially overlapping to Gruber's criteria: *Consistency* and *Expendability*, which are similar to Gruber's coherence and extendibility respectively; *Conciseness* – definitions should be clear and unambiguous and yet expressed in few words; *Completeness* – the ontology captures all that is known about the real world in a finite structure; and *Sensitiveness* – how sensitive the ontology is to small changes in a given definition.

Some criteria (e.g. expendability, clarity and completeness) are difficult to evaluate and require manual (and subjective) inspection by domain experts or ontology engineers (Yu *et al.* 2007). Other criteria can be measured quantitatively by various measures. For example, consistency can be measured based on the number of circularity errors, partition errors and semantic inconsistency errors, and conciseness can be measured based on the number of redundancy error, grammatical redundancy errors and number of identical formal definition of classes (Gómez-Pérez 2001).

Selection of appropriate criteria and measures for the evaluation an ontology depends on the ontology requirements and goals, as demonstrated by Yu et al. (2007).

- *Data driven evaluation* – In this evaluation method the ontology is compared with relevant sources of data (e.g. documents, dictionaries, etc.) about the domain of discourse. For example, Brewster et al. (2004) uses such a method to determine the degree of structural fit between an ontology and a corpus of documents.

To the above classification of methods, Brank et al. (2005) add another dimension of classification – one that is based on the level of evaluation. They argue that an ontology could be evaluated on six different levels, representing different ontology aspects, as follows. First, the *lexical, vocabulary, or data* level – in this level the concepts, facts or instances that are included in the ontology are evaluated, usually by comparing them with different domain-related data-sources, as well as string similarities techniques (e.g. in Maedche and Staab 2002). Second, the *hierarchy or taxonomy* level – in this level we evaluate whether an appropriate hierarchy of 'is-a' relations between concepts is defined (e.g. Yu *et al.* 2007). Third, *other semantic relations* – in this level, semantic relations, besides 'is-a', are evaluated. Forth, the *context* level – an ontology may be part of a larger collection of ontologies and may reference or be referenced by definitions in these other ontologies. In this level we evaluate these inter-ontology references. The context of an ontology could also be an application that uses the ontology. If this is the case, we evaluate how effective the ontology is with respect to the achieving the application goals. Fifth, the *syntactic* level – here we evaluate whether the ontology is correctly specified and follows the syntactic requirements of the selected formal language. This level is relevant when the ontology is created manually. Many of the existing ontology editors (e.g. Protégé) include automatic mechanisms for syntactic evaluation. Finally, the structure, architecture and design level – in this level we evaluate whether the ontology satisfies pre-defined design principles or criteria, structural concerns and whether it is suitable for further development.

Brank et al. (2005) map between the four evaluation methods (i.e., gold standard, application-based, criteria-based and data-driven) and the above levels. For example, while all methods are suitable for evaluating the 'lexical, vocabulary, or data' level, the 'hierarchy or taxonomy' level, and the 'other semantic relations' level, only the application-based and criteria-based methods are suitable for evaluating the 'context' level. In addition, only the golden-standard and criteria-based methods are suitable for evaluating the 'syntactic' level, and only the criteria-based method is suitable for evaluating the 'structure, architecture and design' level.

*Ontology Purposes in Agriculture*

The selection of appropriate evaluation methods for a given ontology should account for the ontology's purpose. Our literature review, reveals that in the context of agriculture, ontologies are used for four main purposes: as means to *share vocabularies and integrate data* (e.g. Chang *et al.* 2008), for *knowledge search and exploration* (e.g. Beck *et al.* 2005; Chang *et al.* 2008; Li *et al.* 2013; Palavitsinis and Manouselis 2014; Song *et al.* 2012), for *system interoperability* (e.g. Aqeel-ur and Zubair 2011; Liao *et al.* 2015; Tomic *et al.* 2015), and for *decision support and for automating decisions* (e.g. Bournaris and Papathanasiou 2012; Gaire *et al.* 2013). The evaluation of ontologies with these different purposes requires focusing on different ontology aspects, i.e., calls for evaluation of different ontology levels (Brank *et al.* 2005), and thus requires using different evaluation methods.

**Results**

*Proposed framework*

To facilitate the selection of suitable evaluation methods, we propose a framework that links between the ontology purpose, the ontology aspects (levels), and the appropriate evaluation method (Brank *et al.* 2005; Yu *et al.* 2007). The framework recommends appropriate evaluation methods based on the purpose of the ontology.

For example, evaluation of the hierarchy or taxonomy level is in particular important in ontologies that are aimed at supporting knowledge search and exploration in order to foster efficient search of knowledge. This can be done using any of the evaluation methods (Brank *et al.* 2005), but particularly suitable are the gold standard method (assuming a gold standard exists) and the criteria based method (e.g. Guarino and Welty 2002). The evaluation of the context level is important in particular in ontologies that are aimed at decision support –for measuring the effectiveness of the ontology-based DSS. In addition, ontologies aimed at integration of data from different ontologies may also benefit from the evaluation of the context level – in order to ensure that references between ontologies are correct. This can be attained using an application-based method or using experts' assessments of predefined criteria (Brank *et al.* 2005).

Evaluation of the semantic relations level is in particular important in ontologies that are aimed at creating shared vocabularies and integrating data (or ontologies) to create system interoperability, since they require agreement on the meaning of things. Such evaluation can be attained by each one of the evaluation methods. Syntactic level evaluation is important for any ontology, as correct syntactic specification is mandatory in order that the ontology can be used by information systems, computerized agents and the Semantic Web. Likewise, since the building blocks of any ontology are the concepts, instances or facts that form it, evaluation of the lexical level, i.e., the vocabulary used to represent these concepts and facts, is important for any ontology. Evaluation on this level usually involve data-driven evaluation methods.

The framework recommendations are summarized in **Fig. 1**. Each quarter of **Fig. 1** links a particular ontology purpose to ontology levels that are important for evaluation, which are linked to suitable evaluation methods. The links to and from each level are marked with a unique colour and line type. When several methods are suitable, a thicker link to a method indicates that the method is more suitable.

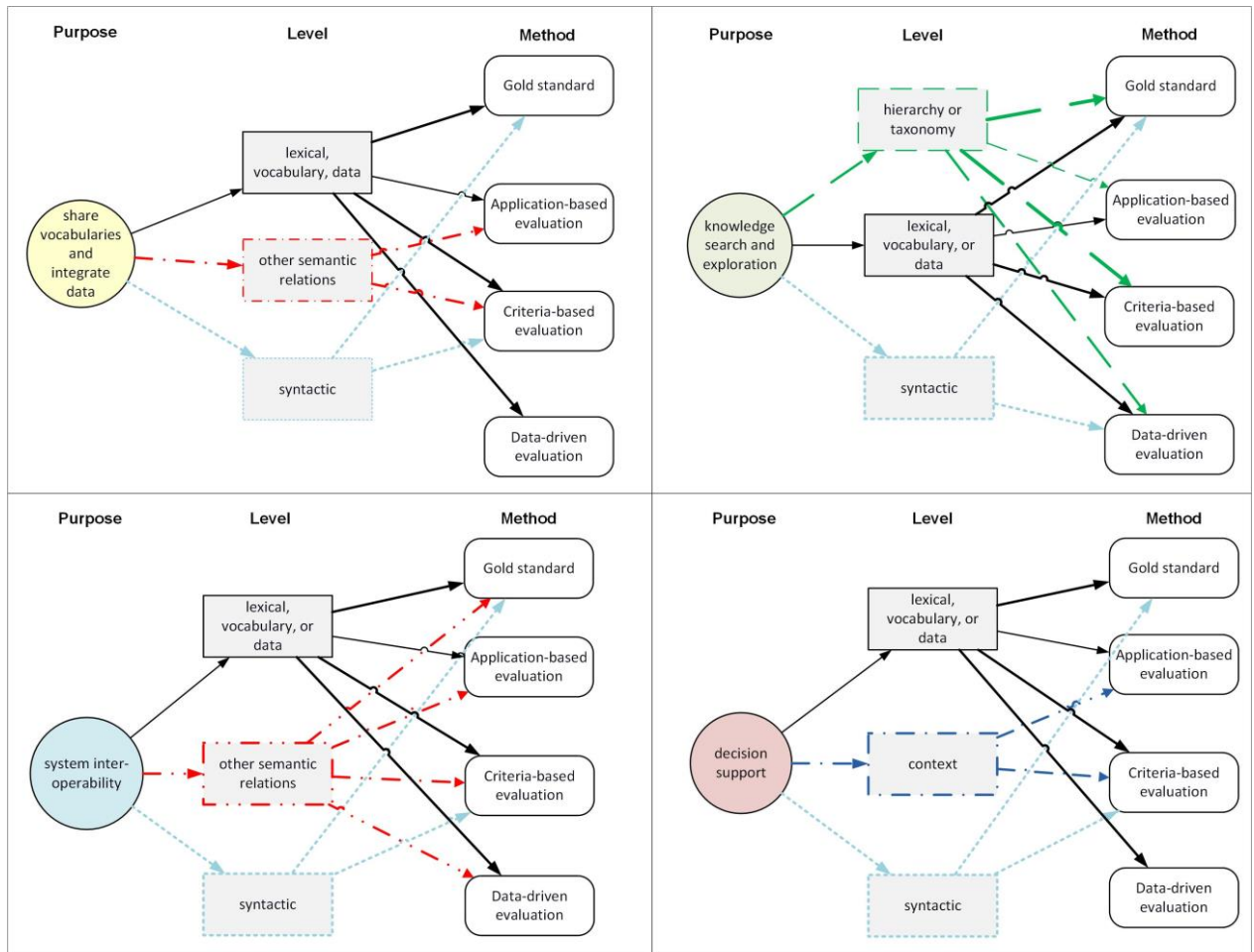

**Fig. 1 A framework for evaluation method selection**

To complement the framework, we propose an iterative evaluation process, in which, we identify the ontology purposes, for each purpose, different ontology levels are selected for evaluation and for each level suitable evaluation methods are applied. The process is depicted in **Error! Reference source not found.**.

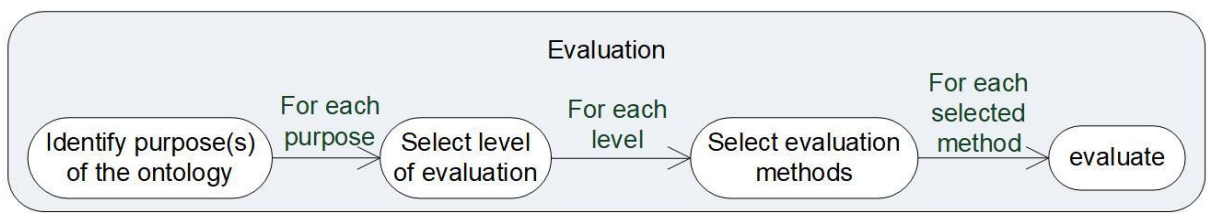

**Fig. 2**. **Ontology evaluation method**

*Framework Application in the Case of a Pest-Control Ontology*

Step 1: purpose identification

The proposed pest-control ontology is intended to serve two purposes: 1. integration of knowledge from different websites as well as concepts from other ontologies in order to provide a pest-control knowledge base, and 2. supporting pest-control decisions.

Step 2: Selecting levels of evaluation

The integrative role of the ontology requires evaluating the *lexical level* (to ensure that the vocabulary used by the ontology is sufficient), the *semantic relations level* (to ensure that there are no semantic ambiguities among concepts from different sources), and the *syntactic level,* as shown in the upper left side of Fig. 1. The decision support role of the ontology requires, in addition to the lexical and syntactic levels, the evaluation of the *context level*.

While the hierarchy level may also be important for evaluation in ontologies aimed at facilitating knowledge search, since our ontology does not specify crops and pests' hierarchies (e.g. categorization of crops to families) but relies on the hierarchies of AGROVOC, we do not evaluate the hierarchy level.

Step 3: Selecting evaluation methods

For each level of evaluation, corresponding evaluation methods are selected.

The evaluation of the *syntactic level* is easily obtained: Protégé (Noy and McGuinness 2001), our ontology editor, provides automatic syntax checking for OWL ontologies. Protégé also provides mechanisms that not only assure correct syntax but also assert that the ontology is free from logical problems.

To evaluate the *Semantic relations* level, *Criteria-based* evaluation is often used (e.g., Delir Haghighi *et al.* 2013; Yu *et al.* 2007). Among the criteria specified above in the Materials and Methods section, we consider the following as relevant for evaluating the semantic relations level: clarity, coherence, minimal encoding bias, conciseness, and completeness (other criteria such as extendibility and minimal ontological commitment, are more related to the context level). While a gold-standard approach may also be applicable, to the best of our knowledge, there are no relevant gold standards or other pest-control ontologies available.

To evaluate the *Context* level, we examine the usability of the ontology and the effectiveness of the system that uses the ontology (Brank *et al.* 2005). In our context, the goals of evaluation are: 1. Validate the usability of the ontology for supporting pesticide usage decisions; 2. Evaluate how using the ontology-based application improves users' performance.

To address the first goal, we developed a prototypical Web-based application for supporting pesticide-usage decisions that is based on the proposed pest-control ontology. To evaluate the effectiveness of the ontology-based application we performed an experiment, in which participants were given two simple tasks which required retrieving information on pesticide application regulation. The experiments measure the increased effectiveness of using the Web-based application in accomplishing those tasks.

To evaluate the lexical level, comparisons with various sources of data are usually used (Brank *et al.* 2005). Such evaluation would be redundant in our case, as the ontology was build based on existing online data sources (see the Specify Formal Ontology section).

Step 4: evaluation

A comprehensive demonstration of the application of the above mentioned evaluation methods in the context of the pest-control ontology is detailed in Goldstein et al. 2019.

**Discussion**

Ontologies are a powerful tool for representing domain knowledge, and thus a growing number of knowledge management systems and DSS are based on ontologies, among other things, in the context of agriculture.

In this paper, a framework for ontology evaluation is presented. The framework combines ideas from pivotal existing evaluation methods (Brank *et al.* 2005; Yu *et al.* 2007). Its use is then demonstrated on the case of pest-control ontology.

Clear and structured methods for ontology evaluation are highly important – without them, it is difficult to consider an ontology as a contribution to research and practice. With the growing use of ontologies and the Semantic Web for developing agricultural systems, such methods become increasingly important. Furthermore, as revealed by the literature review, most of the studies that develop agricultural ontologies do not present the method of development and, even worse, do not discuss how the developed ontologies were evaluated, making it difficult to share and reuse them. Our work thus narrows this gap in the literature of agricultural ontologies, by proposing a development and evaluation method that can be followed by other studies.